\documentclass{article}

\PassOptionsToPackage{numbers, compress}{natbib}

\usepackage[preprint]{neurips_2026}
\usepackage{graphicx}
\usepackage{pifont}
\usepackage{amsmath}

\usepackage[utf8]{inputenc} 
\usepackage[T1]{fontenc}    
\usepackage{hyperref}       
\usepackage{url}            
\usepackage{booktabs}       
\usepackage{amsfonts}       
\usepackage{nicefrac}       
\usepackage{microtype}      
\usepackage{xcolor}         
\usepackage{csquotes}
\setcitestyle{numbers, square}
\usepackage{enumitem}
\usepackage{tikz}  \usetikzlibrary{arrows.meta,positioning,fit,backgrounds,calc}

\usepackage{tabularx}
\usepackage{booktabs}
\usepackage{adjustbox}
\usepackage{multirow}
\title{GenSyn10: A Multi-Generative AI Dataset For Benchmarking Image Classification 
}

\author{%
  Md Faraz Kabir Khan \quad Saeed Anwar \quad Ghulam Mubashar Hassan \\
  Department of Computer Science and Software Engineering \\
  The University of Western Australia \\
  Perth, WA 6009 \\
  \texttt{24427672@student.uwa.edu.au, \{saeed.anwar, ghulam.hassan\}@uwa.edu.au}
}

\begin{document}

\maketitle

\begin{abstract}
The rapid advancement of generative AI has outpaced our ability to reliably detect its outputs, particularly when detectors encounter generators they have not seen before. We introduce GenSyn10, a CIFAR-10-aligned synthetic image dataset of 60,000 images (10 classes, 32$\times$32, 50k/10k split) generated using three architecturally diverse state-of-the-art models: FLUX.2-dev (Rectified Flow Transformer), HunyuanImage-3.0 (MoE Transformer), and Qwen-Image-2512 (Multimodal Diffusion Transformer), to advance research in AI-generated image detection. A central challenge in this domain is that detectors perform well on known generators but degrade on unseen ones. GenSyn10 addresses this limitation by curating data from multiple contemporary architectures under a standardized generation protocol, enabling controlled and systematic evaluation of out-of-distribution (OOD) generalization to novel generators. Images are generated using a template-based prompt engine and downsampled to ensure consistency. We evaluate 17 image classification models under a four-stage protocol: real-data baseline, zero-shot transfer, fine-tuning, and retention. Despite a measurable domain gap, CIFAR-10-trained models achieve up to 96.86\% zero-shot accuracy on GenSyn10, increasing to 99.88\% after fine-tuning. In binary real-vs-synthetic classification, fine-tuned models achieve 97-99.9\% accuracy on seen generators but drop to 79-96\% on images from an unseen generator, highlighting persistent limitations in OOD generalization. These results establish GenSyn10 as a controlled benchmark for studying synthetic image detection beyond single-generator settings, supporting research on robustness, domain adaptation, and cross-generator generalization.

The datasets, codes, and trained model weights are available at \url{https://github.com/farazkabir/GenSyn10-Benchmarking}
\end{abstract}

\section{Introduction}
\label{sec:intro}

Recent advancements in generative modeling have enabled the rapid synthesis of highly realistic images from textual descriptions in seconds. Contemporary text-to-image systems increasingly combine powerful generative frameworks with multimodal or vision-language components, thereby enhancing semantic conditioning and precise adherence to prompts. Nonetheless, generated images continue to exhibit subtle yet systematic discrepancies relative to real-world images, attributable to learned biases, semantic gaps, distributional constraints, and architecture-specific artifacts. A comprehensive understanding of these differences is vital for research on generative realism, distributional shifts, model biases, and the dependable detection of AI-generated images.

Early Generative Adversarial Networks (GANs)~\cite{goodfellow2014generative,karras2019style,karras2018progressive} demonstrated impressive synthesis capabilities but were constrained by instability, mode collapse, and limited diversity.
Diffusion and latent diffusion models~\cite{ho2020denoising,rombach2022high} later became dominant, powering systems such as Stable Diffusion~\cite{rombach2022high,esser2024scaling} and DALL-E~\cite{ramesh2022dalle2}.
More recently, the field has diversified beyond conventional diffusion.
FLUX~\cite{flux2024} uses rectified flow transformers; Qwen-Image-2512~\cite{qwen2025image} pairs a Multimodal Diffusion Transformer (MMDiT) with a frozen Qwen2.5-VL vision-language model for semantic encoding; and HunyuanImage-3.0~\cite{cao2025hunyuanimage} adopts an autoregressive Mixture-of-Experts architecture.
These models represent distinct generative families and are likely to produce different artifact signatures. This rapid architectural progress creates two related but distinct needs for the research community.

Detection models are only as current as their training data. A detector trained exclusively on Stable Diffusion images, as in CIFAKE~\cite{bird2024cifake}, has never encountered artifacts from rectified flow transformers, MoE architectures, or VLM-conditioned diffusion models. Publicly available datasets reflecting contemporary generators are needed for training and evaluating detection methods relevant to the current landscape.
Similarly, detectors trained on images from one generative architecture tend to overfit to that architecture's artifacts and fail on outputs from unseen generators~\cite{wang2020cnn,gragnaniello2021gan,corvi2023detection}. This \emph{out-of-distribution (OOD) generator shift} problem is arguably the central open challenge in AI-generated image detection. A benchmark of contemporary generator images would enable direct OOD evaluation, revealing whether existing detectors generalize beyond their training generators. 

To address these needs, we introduce GenSyn10, a CIFAR-10-structured synthetic dataset of 60{,}000 images generated from three modern open-source models: FLUX.2-dev~\cite{flux2-2025}, HunyuanImage-3.0~\cite{cao2025hunyuanimage}, and Qwen-Image-2512~\cite{qwen2025image}. We adopt the CIFAR-10~\cite{cifar10} structure, with ten balanced classes of low-resolution images, because its fixed image size, uniform class distribution, and modest scale minimize confounding factors and enable controlled and reproducible comparisons across models and detection methods. These properties have made CIFAR-10~\cite{cifar10} a recurring testbed for studies on classification, robustness, and synthetic-image detection, thereby ensuring that results on GenSyn10 can be directly contextualized within prior work.

GenSyn10 serves a dual purpose. First, it acts as a training and benchmarking resource: to the best of our knowledge, it is the first CIFAR-10-equivalent dataset of synthetic images from the latest generation of multiple open-source models, allowing researchers to build classifiers, study domain adaptation, or develop detection methods on current-generation data. Second, it acts as an OOD evaluation resource: because its images come from generators absent in prior benchmarks, detectors trained on older datasets can be evaluated directly to assess their robustness to novel generator architectures. To support controlled within-benchmark OOD experiments, we additionally generate images from a held-out fourth generator, Stable Diffusion 3.5 Large~\cite{sd35}, using the identical prompt grammar and downsampling pipeline but excluding them from the GenSyn10 training set.

To demonstrate GenSyn10's utility, we design a four-stage evaluation protocol (baseline $\rightarrow$ zero-shot $\rightarrow$ fine-tuning $\rightarrow$ retention) and apply it to 17 architectures spanning seven model families, from lightweight CNNs (MobileNetV2, MNASNet) to large-scale vision transformers (ViT-Base/16, Swin-Base). This breadth enables conclusions not only about which models perform best, but about which architectural properties confer resilience to OOD generator shift.

Our contributions are:
\begin{enumerate}[nosep, itemsep=2pt,leftmargin=2.75em]
  \item A modern multi-generator synthetic benchmark: a CIFAR-10-aligned dataset of 60{,}000 images from three state-of-the-art open-source generators spanning distinct architectural families, the first benchmark to include images from FLUX.2-dev~\cite{flux2-2025}, HunyuanImage-3.0~\cite{cao2025hunyuanimage}, and Qwen-Image-2512~\cite{qwen2025image} together.
  
  \item A combinatorial prompt grammar: a template-based system capable of producing $>$100{,}000 unique prompt combinations per class with controllable difficulty tiers, ensuring diverse and reproducible generation.
  
  \item A comprehensive classifier benchmark: evaluation of 17 architectures across seven families under a four-stage protocol with family-aware training, establishing a detailed performance baseline on current-generation synthetic images.
  
  \item Systematic OOD diagnosis: binary real-vs-synthetic classification experiments with a held-out generator that quantify the vulnerability of all 17 architectures to unseen generators (roughly 4--18 percentage-point accuracy drops), demonstrating that architectural inductive bias, not model capacity, is the primary determinant of OOD resilience.
\end{enumerate}

\section{Related Work}
\label{sec:related}

\textbf{Evolution of text-to-image generative models.}
Each new generative paradigm introduces distinct artifacts that challenge existing detectors.
The field originated with GANs~\cite{goodfellow2014generative}, whose successors Progressive GAN~\cite{karras2018progressive} and StyleGAN~\cite{karras2019style} improved quality but produced detectable high-frequency artifacts~\cite{wang2020cnn,frank2020leveraging}.
Diffusion and latent diffusion models~\cite{ho2020denoising,rombach2022high} marked a paradigm shift, now underpinning Stable Diffusion~\cite{rombach2022high,esser2024scaling} and DALL-E~\cite{ramesh2022dalle2,betker2023dalle3}. The most recent generation has further diversified.
FLUX.2~\cite{flux2-2025}(released November 2025), the successor to the 12B-parameter FLUX.1~\cite{flux2024}, couples a Mistral-3 24B-parameter vision-language model with a rectified flow transformer~\cite{liu2022flow,esser2024scaling} totalling 32B parameters.
Qwen-Image-2512~\cite{qwen2025image} (December 2025) integrates a frozen 7B-parameter Qwen2.5-VL for semantic encoding, a dual-decoder VAE for image tokenization, and a 20B-parameter MMDiT backbone, replacing CLIP-based text conditioning with a full vision-language model.
HunyuanImage-3.0~\cite{cao2025hunyuanimage} (September 2025) departs from diffusion-transformer designs entirely: its 80B-parameter MoE backbone (13B activated per token, 64 experts) uses autoregressive next-token prediction for text and diffusion-based prediction for image tokens within a unified multimodal framework. Because each family produces fundamentally different artifact signatures, a benchmark must span genuinely distinct architectures.

\textbf{AI-generated image detection.}
Wang et al.~\cite{wang2020cnn} showed that a ProGAN-trained classifier could generalize across GAN architectures with careful augmentation, though this transferability degrades across generator families~\cite{gragnaniello2021gan,frank2020leveraging}.
Similarly, Corvi et al.~\cite{corvi2023detection} confirmed that the OOD generator problem extends across paradigm boundaries: GAN-trained detectors fail on diffusion outputs and vice versa.
Subsequent work pursued more generalizable representations via CLIP features~\cite{ojha2023towards,cozzolino2024raising}, autoencoder reconstruction error~\cite{ricker2024aeroblade}, hybrid CLIP-VAE features~\cite{cheng2025cospy}, and hierarchical classification~\cite{namani2025deepguard}.
A persistent finding is that detectors tend to overfit to generator-specific artifacts that do not generalize to architecturally novel models~\cite{corvi2023detection,hong2025wildfake}.

\textbf{Synthetic image benchmarks.}
The benchmark closest to GenSyn10 is CIFAKE~\cite{bird2024cifake}, which provides 120{,}000 CIFAR-10-aligned $32\!\times\!32$ images (60{,}000 fake) but draws all synthetic samples from a single outdated generator (Stable Diffusion v1.4~\cite{rombach2022high}), preventing cross-generator evaluation. Larger benchmarks present other trade-offs: GenImage~\cite{zhu2024genimage} (2.68M images, eight generators) mixes heterogeneous resolutions and prompts; WildFake~\cite{hong2025wildfake} (3.57M images) reports that ViTs outperform ResNet-50 in cross-generator settings but is confounded by resolution and compression artifacts; DeFake~\cite{sha2023defake} lacks standardized generation conditions; and CO-SPYBENCH~\cite{cheng2025cospy} and DeepGuardDB~\cite{namani2025deepguard} are either framework-specific or small in scale.
Three limitations recur across these benchmarks: none include images from the most recent open-source generators (FLUX.2~\cite{flux2-2025}, Qwen-Image-2512~\cite{qwen2025image}, HunyuanImage-3.0~\cite{cao2025hunyuanimage}), most lack standardized prompt and resolution conditions, and none provide a held-out generator protocol for within-benchmark out-of-distribution (OOD) evaluation. GenSyn10 directly addresses all three.

\textbf{Vision architectures for image classification.}
Our benchmark spans VGG~\cite{simonyan2014very}, ResNet~\cite{he2016deep}, DenseNet~\cite{huang2017densely}, GoogLeNet/Inception~\cite{szegedy2015going,szegedy2016rethinking}, MobileNetV2~\cite{sandler2018mobilenetv2}, MNASNet~\cite{tan2019mnasnet}, EfficientNet~\cite{tan2019efficientnet,tan2021efficientnetv2}, RegNet~\cite{radosavovic2020designing}, ViT~\cite{dosovitskiy2021an}, Swin Transformer~\cite{liu2021swin}, ConvNeXt~\cite{liu2022convnet}, and ConvNeXtV2~\cite{woo2023convnextv2}.
Prior benchmarks evaluated far fewer detectors~\cite{hong2025wildfake,zhu2024genimage,bird2024cifake,namani2025deepguard}.
Our 17-model evaluation across seven families enables analysis at the level of architectural \emph{properties} (local convolutions, global self-attention, hybrid designs) rather than at the level of individual models.

\begin{figure}[t]
   
    \centering
    \setlength{\tabcolsep}{2pt} 
    \renewcommand{\arraystretch}{0.9} 

    \resizebox{\linewidth}{!}{%
    \begin{tabular}{@{}c *{8}{c}@{}}
               & \multicolumn{2}{c}{FLUX}  & \multicolumn{2}{c}{QWEN} & \multicolumn{2}{c}{Hunyuan} & \multicolumn{2}{c}{Stable Diffusion}\\
      Airplane &
      \includegraphics[width=.09\textwidth]{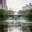} &
      \includegraphics[width=.09\textwidth]{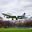} &
      \includegraphics[width=.09\textwidth]{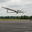} &
      \includegraphics[width=.09\textwidth]{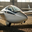} &
      \includegraphics[width=.09\textwidth]{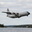} &
      \includegraphics[width=.09\textwidth]{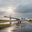} &
      \includegraphics[width=.09\textwidth]{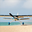} &
      \includegraphics[width=.09\textwidth]{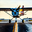} \\[-2pt]

      Automobile &
      \includegraphics[width=.09\textwidth]{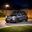} &
      \includegraphics[width=.09\textwidth]{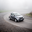} &
      \includegraphics[width=.09\textwidth]{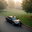} &
      \includegraphics[width=.09\textwidth]{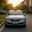} &
      \includegraphics[width=.09\textwidth]{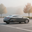} &
      \includegraphics[width=.09\textwidth]{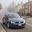} &
      \includegraphics[width=.09\textwidth]{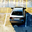} &
      \includegraphics[width=.09\textwidth]{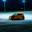} \\[-2pt]

      Bird &
      \includegraphics[width=.09\textwidth]{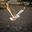} &
      \includegraphics[width=.09\textwidth]{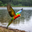} &
      \includegraphics[width=.09\textwidth]{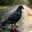} &
      \includegraphics[width=.09\textwidth]{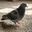} &
      \includegraphics[width=.09\textwidth]{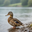} &
      \includegraphics[width=.09\textwidth]{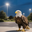} &
      \includegraphics[width=.09\textwidth]{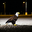} &
      \includegraphics[width=.09\textwidth]{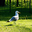} \\[-2pt]

      Cat &
      \includegraphics[width=.09\textwidth]{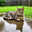} &
      \includegraphics[width=.09\textwidth]{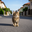} &
      \includegraphics[width=.09\textwidth]{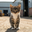} &
      \includegraphics[width=.09\textwidth]{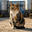} &
      \includegraphics[width=.09\textwidth]{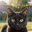} &
      \includegraphics[width=.09\textwidth]{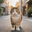} &
      \includegraphics[width=.09\textwidth]{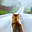} &
      \includegraphics[width=.09\textwidth]{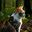} \\[-2pt]

      Deer &
      \includegraphics[width=.09\textwidth]{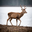} &
      \includegraphics[width=.09\textwidth]{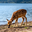} &
      \includegraphics[width=.09\textwidth]{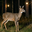} &
      \includegraphics[width=.09\textwidth]{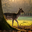} &
      \includegraphics[width=.09\textwidth]{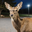} &
      \includegraphics[width=.09\textwidth]{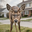} &
      \includegraphics[width=.09\textwidth]{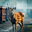} &
      \includegraphics[width=.09\textwidth]{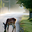} \\[-2pt]

      Dog &
      \includegraphics[width=.09\textwidth]{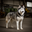} &
      \includegraphics[width=.09\textwidth]{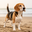} &
      \includegraphics[width=.09\textwidth]{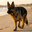} &
      \includegraphics[width=.09\textwidth]{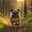} &
      \includegraphics[width=.09\textwidth]{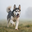} &
      \includegraphics[width=.09\textwidth]{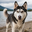} &
      \includegraphics[width=.09\textwidth]{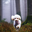} &
      \includegraphics[width=.09\textwidth]{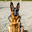} \\[-2pt]

      Frog &
      \includegraphics[width=.09\textwidth]{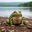} &
      \includegraphics[width=.09\textwidth]{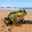} &
      \includegraphics[width=.09\textwidth]{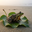} &
      \includegraphics[width=.09\textwidth]{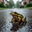} &
      \includegraphics[width=.09\textwidth]{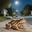} &
      \includegraphics[width=.09\textwidth]{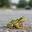} &
      \includegraphics[width=.09\textwidth]{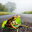} &
      \includegraphics[width=.09\textwidth]{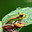} \\[-2pt]

      Horse &
      \includegraphics[width=.09\textwidth]{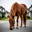} &
      \includegraphics[width=.09\textwidth]{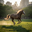} &
      \includegraphics[width=.09\textwidth]{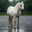} &
      \includegraphics[width=.09\textwidth]{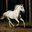} &
      \includegraphics[width=.09\textwidth]{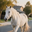} &
      \includegraphics[width=.09\textwidth]{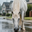} &
      \includegraphics[width=.09\textwidth]{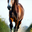} &
      \includegraphics[width=.09\textwidth]{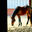} \\[-2pt]

      Ship &
      \includegraphics[width=.09\textwidth]{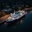} &
      \includegraphics[width=.09\textwidth]{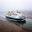} &
      \includegraphics[width=.09\textwidth]{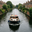} &
      \includegraphics[width=.09\textwidth]{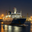} &
      \includegraphics[width=.09\textwidth]{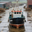} &
      \includegraphics[width=.09\textwidth]{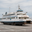} &
      \includegraphics[width=.09\textwidth]{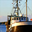} &
      \includegraphics[width=.09\textwidth]{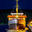} \\[-2pt]

      Truck &
      \includegraphics[width=.09\textwidth]{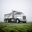} &
      \includegraphics[width=.09\textwidth]{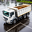} &
      \includegraphics[width=.09\textwidth]{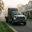} &
      \includegraphics[width=.09\textwidth]{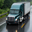} &
      \includegraphics[width=.09\textwidth]{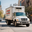} &
      \includegraphics[width=.09\textwidth]{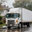} &
      \includegraphics[width=.09\textwidth]{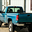} &
      \includegraphics[width=.09\textwidth]{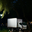}
    \end{tabular}%
    }

    \caption{Diverse samples from the GenSyn10 dataset illustrating variations within each of its 10 object categories.}
    \label{fig:samples}
\end{figure}

\section{Dataset Construction}
\label{sec:dataset}

The GenSyn10 pipeline comprises three stages: prompt generation, image synthesis, and post-processing.

\subsection{Prompt Generation}
\label{sec:prompts}
We design a \emph{combinatorial prompt grammar} that generates diverse, photorealistic prompts balanced across the 10 classes of CIFAR-10. Each prompt is assembled from six independently sampled slots: (i) Instance, per-class subtypes (6–7 per class); (ii) Scene, background setting (14 options); (iii) View, camera angle (8 standard, 10 for hard tier); (iv) Lighting, illumination condition (8 options); (v) Action, class-specific pose (4–6 per class); and (vi) Prompt opener, introductory phrase for linguistic variety (6 options). Prompts follow the template:

\begin{quote}
\texttt{\{opener\} of a \{instance\} \{class\} \{action\} in \{scene\}, \{view\}, \{lighting\}. [hard\_modifier.] \{CONSTRAINTS\}.}
\end{quote}

\noindent where \texttt{CONSTRAINTS} is a fixed suffix enforcing photorealism (single main subject, natural colors, realistic texture, sharp focus, no people/text/watermarks/logos/borders/collage/CGI/illustration). A global negative prompt is provided for models supporting classifier-free guidance. We define two tiers: Clean (80\% of prompts; standard views, no modifiers) and Hard (20\%; extended views such as \enquote{cropped at the edge of the frame}, plus a random hard modifier such as \enquote{partially occluded} or \enquote{slight motion blur}). The combinatorial space exceeds 100{,}000 unique clean combinations per class, ensuring minimal repetition. We generate 10{,}000 prompts (1{,}000 per class), allowing each prompt to be used twice per generator.

\subsection{Image Generation} 
\label{sec:generation}

We select three state-of-the-art open-source text-to-image models representing distinct architectural paradigms, each among the highest-performing models currently available for reproducible research. All models share a common generation pipeline:

\begin{itemize}
\item Use three open-source generators with publicly available weights spanning distinct architectural paradigms: FLUX.2-dev (4-bit BnB), HunyuanImage-3.0 (NF4 4-bit), and Qwen-Image-2512 (bfloat16). In addition, we reserve Stable Diffusion 3.5 Large (bfloat16) as a held-out generator, not used to construct GenSyn10, enabling evaluation of generalization to an unseen synthetic distribution.
\item Load per-class prompt files and sample $N$ prompts per class ($N=1000$).
\item Generate images at $512\!\times\!512$ resolution using model-specific inference.
\item Downsample to $32\!\times\!32$ via Lanczos resampling.
\item Save both high-resolution and CIFAR-sized images with per-image metadata (prompt, seed, class, timestamp, model ID).  Image filenames follow the format \texttt{\{prompt\_id\}\_\{seed\}.png}.
\end{itemize}

Deterministic seeding ensures reproducibility: each image's random seed is derived from a hash of the global seed, the model backend, the class label, the prompt ID, and the prompt text. The pipeline supports resumption: if output files already exist for a given task, generation is skipped.

\paragraph{Design rationale.}
Selecting generators from three distinct architectural families supports both of GenSyn10's core objectives. As a \emph{training resource}, it exposes models to artifact signatures from rectified flow (FLUX.2-dev~\cite{flux2-2025}), autoregressive MoE (HunyuanImage-3.0~\cite{cao2025hunyuanimage}), and an MMDiT architecture with Qwen2.5-VL conditioning (Qwen-Image-2512~\cite{qwen2025image})---three paradigms absent from existing benchmarks. Released in late 2025, these models rank among the top open-source entries on the LM Arena Text-to-Image leaderboard~\cite{chiang2024chatbot} as of early 2026. As an \emph{OOD evaluation} resource, it enables detectors trained on older datasets to be tested against novel artifact signatures, providing a rigorous measure of cross-generator generalization.

Restricting selection to open-source models also avoids copyright issues and aids reproducibility. We use quantized weights for generation since our low target resolution makes full-precision inference unnecessary for visual fidelity. Stable Diffusion 3.5 Large~\cite{sd35} is excluded from the main dataset due to lower output quality and redundancy with prior benchmarks, and is retained as a held-out generator for controlled OOD binary detection experiments (Section~\ref{sec:binary}) to test whether detectors genuinely generalize rather than overfit to seen generators.

\subsection{Post-Processing and Dataset Structure}
The 60{,}000 images (20{,}000 per generator, balanced across 10 classes) are merged and split into a training set of 50{,}000 images (5{,}000 per class) and a test set of 10{,}000 images (1{,}000 per class), mirroring the CIFAR-10 split exactly. All images are stored as $32\!\times\!32$ RGB PNGs organized in class-labeled subdirectories. For the binary classification experiments, we source the real samples from the CIFAR-10 dataset and use the generated samples as the synthetic class. Generated image samples are presented in Figure~\ref{fig:samples}.

\subsection{Dataset Quality Assessment}
\label{sec:quality}

We assess the distributional quality of GenSyn10 relative to the real CIFAR-10~\cite{cifar10} training set using established fidelity and diversity metrics. Findings are shown in Table~\ref{tab:quality}.

\begin{table}[t]
\centering
\caption{Distributional quality metrics for GenSyn10 vs.\ real CIFAR-10.}
\label{tab:quality}
\small
\begin{tabular}{@{}lrl@{}}
\toprule
\textbf{Metric} & \textbf{Value} & \textbf{Interpretation} \\
\midrule
Inception Score (IS)~\cite{salimans2016improved} & $8.77 \pm 0.11$ & Class diversity \& image clarity ($\uparrow$) \\
Fr\'{e}chet Inception Distance (FID)~\cite{heusel2017gans} & 35.17 & Distributional similarity ($\downarrow$) \\
Precision~\cite{kynkaanniemi2019improved} & 0.244 & Fidelity: fraction in real manifold ($\uparrow$) \\
Recall~\cite{kynkaanniemi2019improved} & 0.263 & Diversity: real manifold coverage ($\uparrow$) \\
\bottomrule
\end{tabular}
\end{table}

\textbf{Interpretation.}
The IS of 8.77 is competitive with real CIFAR-10~\cite{cifar10} ($\approx$11.2), indicating that GenSyn10 preserves class diversity and image clarity. The FID of 35.17 reflects a moderate domain gap, expected given the different generation process and Lanczos downsampling from $512\!\times\!512$ to $32\!\times\!32$; the relatively low precision (0.244) and recall (0.263) are attributable to smoother synthetic textures versus the natural noise of real CIFAR-10. This gap is a \emph{feature} rather than a limitation: a trivially small gap would make detection uninformative, while an excessively large gap would make it artificially easy. The high zero-shot accuracy of CIFAR-10-trained classifiers (up to 96.86\%, Section~\ref{sec:fourstage}) confirms that the gap preserves semantic content while introducing meaningful distributional differences suitable for both training and evaluation.
\section{Benchmark Protocol and Experiments}
\label{sec:experiments}

\begin{figure}[t]
  \centering
  \includegraphics[width=1\linewidth]{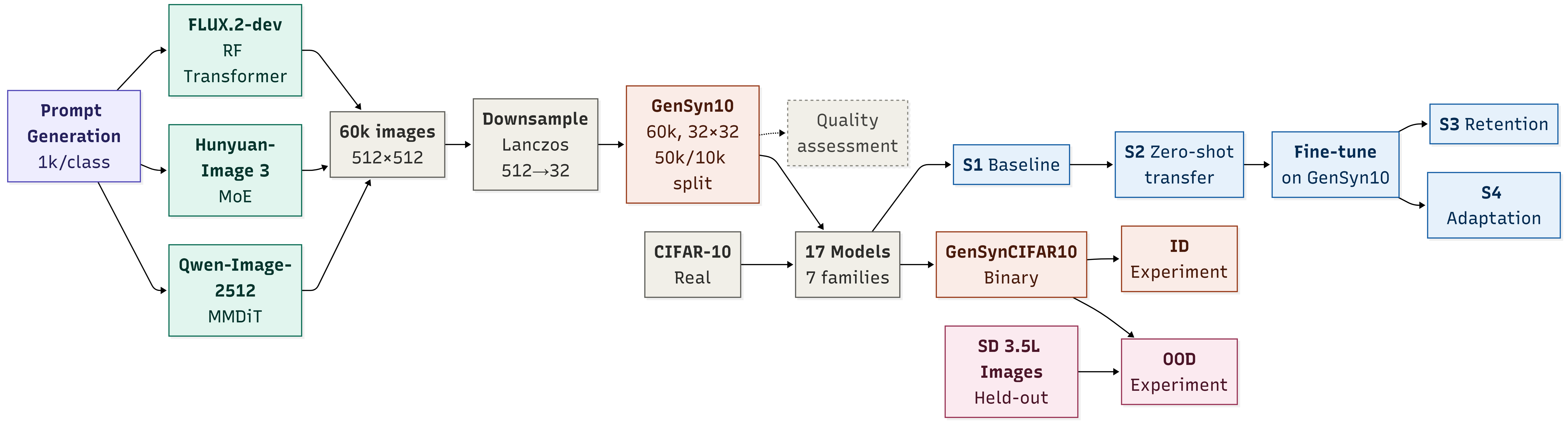}
  \caption{Overview of the benchmark pipeline, including prompt generation, image synthesis using modern generators, GenSyn10 dataset construction, model fine-tuning, and out-of-distribution evaluation using a held-out generator.}
  \label{fig:GenSyn10_pipeline}
\end{figure}

We evaluate 17 diverse classification architectures spanning CNNs and vision transformers (Table~\ref{tab:architectures}). All experiments use the same codebase, hyperparameters, and evaluation pipeline to ensure fair comparison. The overview of our benchmark protocols and experiments is shown in Figure~\ref{fig:GenSyn10_pipeline}.

\subsection{Weight Initialization and Fine-tuning}

Models are initialized using the best available CIFAR-10-compatible weights following a three-tier strategy: CIFAR-10-native checkpoints~\cite{chenyaofo2023cifar,huyvnphan2021cifar10}, CIFAR-10-finetuned checkpoints from the Hugging Face Hub, or ImageNet-1k pretrained weights~\cite{deng2009imagenet} with a family-aware CIFAR-10 fine-tuning protocol for the other architectures. Initialization sources are summarized in Table~\ref{tab:architectures}.
\begin{table}[tbp]
  \centering
  \caption{Classification architectures evaluated in the GenSyn10 benchmark. \emph{Left}: models with CIFAR-10 checkpoints (native or Hugging Face Hub). \emph{Right}: models transfer-learned from ImageNet-1k via family-aware fine-tuning.}
  \label{tab:architectures}
  \small
  \setlength{\tabcolsep}{4pt}
  \begin{tabular}{@{}lll@{\hskip 14pt}ll@{}}
    \toprule
    \multicolumn{3}{c}{\textbf{CIFAR-10 Checkpoints}} & \multicolumn{2}{c}{\textbf{ImageNet-1k Transfer}} \\
    \cmidrule(r){1-3} \cmidrule(l){4-5}
    \textbf{Family} & \textbf{Architecture} & \textbf{Source} & \textbf{Family} & \textbf{Architecture} \\
    \midrule
    VGG~\cite{simonyan2014very}       & VGG19-BN      & Native & ResNet~\cite{he2016deep}               & ResNet-152       \\
    ResNet~\cite{he2016deep}          & ResNet-50      & Native & DenseNet~\cite{huang2017densely}       & DenseNet-201     \\
    DenseNet~\cite{huang2017densely}  & DenseNet-121   & Native & MNASNet~\cite{tan2019mnasnet}          & MNASNet-1.0      \\
                                      & DenseNet-169   & Native & EfficientNet~\cite{tan2019efficientnet}& EfficientNet-B0  \\
    Inception~\cite{szegedy2015going} & GoogLeNet      & Native & \cite{tan2021efficientnetv2}           & EfficientNetV2-S \\
    \cite{szegedy2016rethinking}      & InceptionV3    & Native & RegNet~\cite{radosavovic2020designing} & RegNetY-8GF      \\
    MobileNet~\cite{sandler2018mobilenetv2} & MobileNetV2 & Native & ConvNeXt~\cite{liu2022convnet}      & ConvNeXt-Base    \\
    ViT~\cite{dosovitskiy2021an}      & ViT-Base/16    & HF Hub~\cite{wu2024vit_hf}  & \cite{woo2023convnextv2}  & ConvNeXtV2-Base  \\
    Swin~\cite{liu2021swin}           & Swin-Base      & HF Hub~\cite{liu2024swin_hf}&                           &                  \\
    \bottomrule
  \end{tabular}
\end{table}

All fine-tuning, both for the CIFAR-10 initialization fallback and for GenSyn10 adaptation, follows a two-phase procedure: (i) classifier head warm-up with frozen backbone, then (ii) full-network fine-tuning.
Optimization uses AdamW~\cite{loshchilov2017adamw} with ReduceLROnPlateau scheduling (factor 0.5, scheduler patience 1, minimum LR $10^{-6}$), label smoothing~\cite{muller2019does} ($\epsilon=0.05$), gradient clipping (max norm 1.0), and mixed-precision training. The best checkpoint is selected by a composite score: $0.7 \times \text{val\_macro\_F1} + 0.3 \times \text{val\_balanced\_accuracy}$.


Because larger transformer-based models are more prone to catastrophic forgetting of pretrained features under aggressive learning rates, we adopt \emph{model-family-specific} hyperparameters. The key distinctions for transformer-class models (ViT, Swin) and hybrid architectures (ConvNeXt, ConvNeXtV2) are: (a)~an additional head-only warm-up epoch (3 vs.\ 2) to stabilize the randomly initialized head before unfreezing pretrained weights; (b)~a $2\times$ lower full-network learning rate ($5\!\times\!10^{-5}$ for ViT/Swin vs.\ $10^{-4}$ for generic CNNs), reflecting the sensitivity of self-attention weights to large gradient updates; (c)~$50\%$ more full fine-tuning epochs (12 vs.\ 8) to allow the slower learning rate to converge; and (d)~higher early-stopping patience (4 vs.\ 3) to avoid premature termination during the longer schedule.
MNASNet receives an intermediate configuration with 10 full epochs and a slightly lower learning rate, as its depthwise-separable convolutions are more fragile than standard convolutions but less sensitive than attention layers.

\subsection{Four-Stage Cross-Domain Evaluation}
\label{sec:fourstage}

We design a four-stage protocol to comprehensively measure the domain gap between CIFAR-10 and GenSyn10 and the effect of fine-tuning. 

\textbf{Stage 1} (Pre-FT on CIFAR-10) evaluates CIFAR-10-initialized models on the CIFAR-10 test set as a baseline. \textbf{Stage 2} (Pre-FT on GenSyn10) evaluates the same models on the GenSyn10 test set without synthetic fine-tuning, measuring zero-shot transfer. \textbf{Stage 3} (Post-FT on CIFAR-10) fine-tunes on the GenSyn10 training set and evaluates on the CIFAR-10 test set to assess retention and forgetting. \textbf{Stage 4} (Post-FT on GenSyn10) fine-tunes on GenSyn10 and evaluates on the GenSyn10 test set to measure adaptation.

\textbf{Fine-tuning protocol.}
GenSyn10 fine-tuning uses a stratified 90/10 train/validation split from the GenSyn10 training set.
The family-aware two-phase protocol is applied: head-only warm-up followed by full-network fine-tuning with family-specific hyperparameters, composite-score checkpoint selection, and early stopping.

\subsubsection{Results}

\begin{table}[tbp]
\centering
\caption{Four-stage cross-domain evaluation results.
\textbf{S1}: CIFAR-10 baseline; \textbf{S2}: zero-shot on GenSyn10; \textbf{S3}: post-GenSyn10-FT on CIFAR-10 (retention); \textbf{S4}: post-GenSyn10-FT on GenSyn10 (adaptation).
Acc = accuracy (\%), F1 = macro F1 (\%).
Best values per column are \textbf{bolded}.}
\label{tab:fourstage}
\footnotesize
\setlength{\tabcolsep}{2.5pt}
\begin{tabular}{l| cc |cc |cc |cc}
\toprule
 & \multicolumn{2}{c|}{\textbf{S1} (CIFAR-10)} & \multicolumn{2}{c|}{\textbf{S2} (GenSyn10 0-shot)} & \multicolumn{2}{c|}{\textbf{S3} (CIFAR-10 ret.)} & \multicolumn{2}{c}{\textbf{S4} (GenSyn10 FT)} \\
\cmidrule(lr){2-3} \cmidrule(lr){4-5} \cmidrule(lr){6-7} \cmidrule(lr){8-9}
\textbf{Model} & Acc & F1 & Acc & F1 & Acc & F1 & Acc & F1 \\
\midrule
VGG19-BN        & 93.91 & 93.90 & 85.71 & 85.67 & 74.98 & 75.11 & 98.75 & 98.75 \\
ResNet-50        & 93.66 & 93.66 & 84.73 & 84.66 & 52.02 & 50.72 & 98.73 & 98.73 \\
ResNet-152       & 97.70 & 97.70 & 95.57 & 95.55 & 78.32 & 77.96 & 99.65 & 99.65 \\
DenseNet-121     & 94.06 & 94.05 & 86.39 & 86.35 & 62.55 & 60.82 & 98.59 & 98.59 \\
DenseNet-169     & 94.05 & 94.05 & 85.96 & 85.94 & 64.12 & 63.76 & 98.97 & 98.97 \\
DenseNet-201     & 96.14 & 96.14 & 94.24 & 94.21 & 82.84 & 82.89 & 99.70 & 99.70 \\
GoogLeNet        & 92.83 & 92.83 & 85.31 & 85.26 & 65.79 & 65.43 & 99.08 & 99.08 \\
InceptionV3      & 93.74 & 93.74 & 86.38 & 86.38 & 67.24 & 66.59 & 98.83 & 98.83 \\
MobileNetV2      & 94.05 & 94.05 & 84.99 & 84.93 & 73.40 & 72.87 & 98.92 & 98.92 \\
MNASNet-1.0      & 81.67 & 81.52 & 74.32 & 74.47 & 67.19 & 66.69 & 92.20 & 92.27 \\
EfficientNet-B0  & 96.56 & 96.56 & 92.22 & 92.22 & 80.57 & 80.69 & 99.63 & 99.63 \\
EfficientNetV2-S & 97.72 & 97.72 & 94.70 & 94.68 & 83.84 & 83.90 & 99.81 & 99.81 \\
RegNetY-8GF      & 97.18 & 97.18 & 94.09 & 94.08 & 77.57 & 77.96 & 99.73 & 99.73 \\
ViT-Base/16      & 94.97 & 94.99 & 92.77 & 92.78 & \textbf{95.12} & \textbf{95.13} & 99.85 & 99.85 \\
Swin-Base        & \textbf{98.83} & \textbf{98.83} & \textbf{96.86} & \textbf{96.83} & 94.50 & 94.48 & 99.79 & 99.79 \\
ConvNeXt-Base    & 98.47 & 98.47 & 96.71 & 96.70 & 92.60 & 92.66 & \textbf{99.88} & \textbf{99.88} \\
ConvNeXtV2-Base  & 98.28 & 98.28 & 96.40 & 96.38 & 73.63 & 73.18 & 99.81 & 99.81 \\
\bottomrule
\end{tabular}
\end{table}

\textbf{Key findings.} Table~\ref{tab:fourstage} reports the main results. i) \textit{High zero-shot transfer.} The best CIFAR-10-trained model (Swin-Base) reaches 96.86\% on GenSyn10, and even the weakest (MNASNet-1.0) 74.32\%, confirming that GenSyn10 preserves CIFAR-10's semantic structure and serves as a drop-in benchmark. ii) \textit{Near-perfect adaptation.} After fine-tuning, all models except MNASNet-1.0 exceed 98.5\% (best: ConvNeXt-Base, 99.88\%); MNASNet-1.0 nonetheless gains the most (+17.88 points), suggesting lightweight architectures benefit disproportionately from synthetic data. iii) \textit{Forgetting is architecture-dependent.} Transformers lose little CIFAR-10 accuracy (mean $\approx$2.1 points for ViT/Swin; ViT-Base/16 retains the best real-data performance at 95.12\%), whereas CNNs degrade far more (e.g., $-$41.6 points for ResNet-50). iv) \textit{Consistent hierarchy.} Transformers and modern hybrids generally outperform classical CNNs, and the best model at every stage is transformer-based, supporting the advantage of global self-attention over purely local receptive fields.

\subsection{Binary Real-vs-Synthetic Classification}
\label{sec:binary}

The binary classification experiment evaluates whether classifiers can distinguish real CIFAR-10 images from synthetic GenSyn10 images and whether this ability generalizes to an unseen generator. For this purpose, we construct GenSynCIFAR10, a balanced binary dataset in which \textit{Class 0} represents real CIFAR-10 images and \textit{Class 1} represents synthetic GenSyn10 images. All 17 models are adapted for binary classification by replacing the original 10-class classification head with a 2-class head and are initialized from their CIFAR-10 checkpoints, where available. Binary fine-tuning follows the same family-aware protocol described in Section~\ref{sec:fourstage}. The primary positive class is \texttt{synthetic}. 

To evaluate robustness under generator shift, each detector is tested in two settings. The in-distribution (ID) setting uses the GenSynCIFAR10 test set, in which synthetic images are drawn from the same generator distribution used during training. The out-of-distribution (OOD) setting uses synthetic images from Stable Diffusion 3.5 Large~\cite{sd35}, a held-out generator not included in GenSyn10 training, and we use the same number of real and synthetic samples (10k each). Since the prompt grammar, image resolution, and post-processing pipeline are kept consistent, the OOD experiment isolates the effect of changing the generator architecture. As shown in Table~\ref{tab:binary_combined}, the same five metrics are reported for both ID and OOD evaluation. ID performance is near-saturated across most models, with Swin-Base achieving the highest accuracy and F1 score of 99.94\%. ConvNeXt-Base and ViT-Base/16 also approach saturation, confirming that real-vs-synthetic detection is highly effective when the generator distribution is known. However, performance consistently drops under the held-out Stable Diffusion 3.5 Large evaluation, ranging from $-$3.80 points (ViT-Base/16) to $-$18.15 points (ConvNeXtV2-Base). This confirms that high ID performance does not guarantee OOD robustness, even when prompt grammar, resolution, and processing are controlled.

\begin{table}[tbp]
\centering
\caption{Binary real-vs-synthetic classification under in-distribution (ID) and out-of-distribution (OOD) evaluation. Recall, specificity, and F1 are computed with \texttt{synthetic} as the positive class. $\Delta$Acc denotes the accuracy drop from ID to OOD. Best values are \textbf{bolded}.}
\label{tab:binary_combined}
\scriptsize
\setlength{\tabcolsep}{2.4pt}
\resizebox{\textwidth}{!}{%
\begin{tabular}{@{}l|ccccc| ccccc |c@{}}
\toprule
\textbf{Model}
& \multicolumn{5}{c}{\textbf{ID: GenSynCIFAR10 Test}}
& \multicolumn{5}{c}{\textbf{OOD: Stable Diffusion 3.5 Large}}
& $\boldsymbol{\Delta}$\textbf{Acc} \\
\cmidrule(lr){2-6}
\cmidrule(lr){7-11}
& \textbf{Acc} & \textbf{Rec} & \textbf{Spec} & \textbf{F1} & \textbf{AUC}
& \textbf{Acc} & \textbf{Rec} & \textbf{Spec} & \textbf{F1} & \textbf{AUC}
& \\
\midrule
VGG19-BN         & 99.52 & 99.56 & 99.48 & 99.52 & 0.9997 & 90.75 & 82.01 & 99.48 & 89.86 & 0.9942 & $-$8.77 \\
ResNet-50        & 99.59 & 99.64 & 99.53 & 99.59 & 0.9998 & 89.85 & 80.17 & 99.53 & 88.76 & 0.9841 & $-$9.74 \\
ResNet-152       & 99.77 & 99.58 & 99.96 & 99.77 & 1.0000 & 91.32 & 82.68 & 99.96 & 90.50 & 0.9920 & $-$8.45 \\
DenseNet-121     & 99.50 & 99.28 & 99.72 & 99.50 & 0.9999 & 89.64 & 79.56 & 99.72 & 88.48 & 0.9885 & $-$9.86 \\
DenseNet-169     & 99.33 & 99.38 & 99.27 & 99.33 & 0.9997 & 92.89 & 86.51 & 99.27 & 92.41 & 0.9935 & $-$6.44 \\
DenseNet-201     & 99.69 & 99.43 & 99.96 & 99.69 & 1.0000 & 91.69 & 83.42 & 99.96 & 90.94 & 0.9962 & $-$8.00 \\
GoogLeNet        & 99.78 & 99.86 & 99.69 & 99.78 & 0.9999 & 94.87 & 90.05 & 99.69 & 94.61 & 0.9889 & $-$4.91 \\
InceptionV3      & 99.31 & 99.71 & 98.91 & 99.31 & 0.9996 & 95.36 & 91.80 & 98.91 & 95.18 & 0.9949 & $-$3.95 \\
MobileNetV2      & 99.61 & 99.55 & 99.67 & 99.61 & 0.9997 & 93.23 & 86.79 & 99.67 & 92.76 & 0.9829 & $-$6.38 \\
MNASNet-1.0      & 97.70 & 95.59 & 99.82 & 97.66 & 0.9996 & 85.48 & 71.15 & 99.82 & 83.06 & 0.9949 & $-$12.22 \\
EfficientNet-B0  & 99.07 & 98.17 & 99.97 & 99.06 & 0.9997 & 86.62 & 73.28 & 99.97 & 84.57 & 0.9929 & $-$12.45 \\
EfficientNetV2-S & 99.72 & 99.43 & \textbf{100.00} & 99.71 & 0.9999 & 91.27 & 82.54 & \textbf{100.00} & 90.43 & 0.9885 & $-$8.45 \\
RegNetY-8GF      & 99.59 & 99.21 & 99.98 & 99.59 & 1.0000 & 90.97 & 81.96 & 99.98 & 90.08 & 0.9971 & $-$8.62 \\
ViT-Base/16      & 99.88 & 99.88 & 99.88 & 99.88 & 1.0000 & \textbf{96.08} & \textbf{92.27} & 99.88 & \textbf{95.92} & 0.9978 & \textbf{$-$3.80} \\
Swin-Base        & \textbf{99.94} & \textbf{99.93} & 99.96 & \textbf{99.94} & \textbf{1.0000} & 95.54 & 91.12 & 99.96 & 95.33 & 0.9966 & $-$4.40 \\
ConvNeXt-Base    & 99.91 & 99.87 & 99.95 & 99.91 & 0.9999 & 94.71 & 89.47 & 99.95 & 94.42 & 0.9987 & $-$5.20 \\
ConvNeXtV2-Base  & 97.34 & 94.70 & 99.99 & 97.27 & 1.0000 & 79.19 & 58.38 & 99.99 & 73.72 & \textbf{0.9988} & $-$18.15 \\
\bottomrule
\end{tabular}%
}
\end{table}

\textbf{Key findings.} \textit{i) ID results are near-saturated.} All models exceed 97\% accuracy, with most surpassing 99\%, indicating the binary task is straightforward when training and test distributions share the same generator. \textit{ii) OOD generator shift reduces reliability.} Every model's accuracy declines on Stable Diffusion 3.5 Large, showing that detectors partly rely on generator-specific features learned during training. \textit{iii) ViT-Base/16 is the strongest OOD detector.} It achieves the best OOD accuracy (96.08\%), recall (92.27\%), and F1 (95.92\%), with the smallest accuracy drop ($-$3.80 points), suggesting transformer representations transfer more effectively to unseen generators than convolutional alternatives. \textit{iv) OOD degradation stems from reduced synthetic recall.} Specificity remains high across models, but synthetic recall drops substantially for ConvNeXtV2-Base, MNASNet-1.0, and EfficientNet-B0, indicating these detectors misclassify OOD synthetic images as real. \textit{v) High AUC suggests ranking ability persists.} Most OOD AUC values remain near 1.000 despite drops in thresholded metrics, indicating the failure reflects imperfect threshold transfer rather than a complete loss of separability.

\section{Discussion, Limitations, and Conclusion}
\label{sec:conclusion}

We present GenSyn10, a CIFAR-10-aligned synthetic image dataset and benchmarking framework addressing two gaps in AI-generated image detection research. First, existing benchmarks rely on outdated generators (CIFAKE: Stable Diffusion v1.4) or aggregate heterogeneous resolutions and prompts that confound controlled analysis~\cite{zhu2024genimage,hong2025wildfake,sha2023defake}. GenSyn10 provides 60{,}000 images from three state-of-the-art open-source generators (FLUX.2-dev, HunyuanImage-3.0, and Qwen-Image-2512) under a standardized protocol, and is the first public benchmark that covers these generators jointly. Second, because GenSyn10 contains images from generators absent from prior benchmarks, detectors trained on older datasets can be evaluated directly for generalization, while a held-out fourth generator (Stable Diffusion 3.5 Large) enables controlled within-benchmark OOD experiments.

Our evaluation of 17 architectures across seven families yields four core findings: \textit{(i)}~CIFAR-10-trained models reach up to 96.86\% zero-shot accuracy on GenSyn10, confirming preserved class semantics; \textit{(ii)}~fine-tuning achieves up to 99.88\% synthetic classification accuracy; \textit{(iii)}~detectors degrade by roughly 4--18 points on the held-out generator, exposing persistent OOD vulnerability; and \textit{(iv)}~transformer-based and hybrid models generally outperform classical CNNs under cross-generator evaluation, suggesting that global receptive fields and self-attention improve resilience to generator shift, while increasing CNN depth or parameter count alone does not guarantee stronger OOD robustness.

\textbf{Controlled prompt diversity.} The combinatorial prompt grammar supports over 100,000 combinations per class, with difficulty tiers varying in viewpoint, occlusion, blur, and composition. Because identical prompts are used across all generators, performance differences can be attributed to generator architecture rather than prompt content, offering stronger experimental control than heterogeneous prior benchmarks~\cite{zhu2024genimage,hong2025wildfake,sha2023defake}.

\textbf{Extensible design.} New generators can be added using the same prompt grammar and downsampling protocol, and the held-out-generator protocol can be expanded with additional unseen models, keeping the benchmark relevant as generative models evolve.

\paragraph{Limitations.} 
Three training generators, plus one held-out generator, provide meaningful architectural diversity but do not cover all paradigms (e.g., consistency models, GAN-based systems), although the modular pipeline makes adding generators straightforward. The template-based grammar produces diverse but structurally regular prompts; free-form prompts may elicit different artifacts. Manual inspection revealed a small number of generation errors---for example, HunyuanImage occasionally generated bulls instead of bullfrogs; the affected prompts were revised by adding scientific names and regenerated, and a few HunyuanImage and Qwen-Image samples that missed the intended subject were reproduced. Minor residual anomalies may remain due to limitations of manual inspection. The artifacts reflect the state of the art as of early 2026, and detection benchmarks will require periodic updating as generators improve. Finally, a single held-out generator demonstrates the OOD methodology but cannot characterize all possible OOD scenarios; future versions can incorporate additional held-out generators.

Overall, GenSyn10 provides a modern, controlled, and extensible benchmark for evaluating synthetic-image classification and OOD robustness. By combining current open-source generator coverage, standardized dataset construction, and broad multi-architecture evaluation, it supports more principled analysis of detector training effectiveness and cross-generator generalization beyond single-generator benchmarks.

\bibliographystyle{plain}
\bibliography{references}

\section*{Technical appendices and supplementary material}
\label{sec: appendice}
\appendix

\section{Prompt Grammar Details}
\label{app:grammar}

\subsection{Class Instances}

Table~\ref{tab:instances} lists the per-class subtypes (instances) used in the prompt grammar.

\begin{table}[h]
\centering
\caption{Per-class instances used in the prompt grammar.}
\label{tab:instances}
\small
\begin{tabular}{@{}lp{10cm}@{}}
\toprule
\textbf{Class} & \textbf{Instances} \\
\midrule
airplane & passenger jet, small prop airplane, biplane, cargo plane, glider, seaplane \\
automobile & sedan, hatchback, coupe, taxi, sports car, compact car \\
bird & sparrow, pigeon, seagull, duck, parrot, eagle \\
cat & tabby cat, black cat, calico cat, siamese cat, longhair cat, kitten \\
deer & doe, stag with antlers, fawn, deer in winter coat, young stag, adult deer \\
dog & labrador, german shepherd, beagle, bulldog, husky, poodle, mixed-breed dog \\
frog & tree frog, bullfrog, poison dart frog, small green frog, brown frog, frog on a leaf \\
horse & pony, racehorse, draft horse, wild horse, brown horse, white horse \\
ship & cargo ship, ferry, sailboat, fishing boat, cruise ship, small motorboat \\
truck & pickup truck, semi truck, dump truck, delivery truck, utility truck, box truck \\
\bottomrule
\end{tabular}
\end{table}

\subsection{Scenes, Views, Lighting, and Actions}

\paragraph{Scenes (14).}
An urban street, a parking lot, a countryside road, a grassy field, a forest clearing, a backyard, a lakeside shore, a sandy beach, an industrial yard, a suburban neighborhood, a city park, a quiet residential street, a gravel lot, a wet roadway after rain.

\paragraph{Views -- Clean tier (8).}
Side profile, 3/4 view, front view, low angle, high angle, close-up, wide shot, slightly off-center framing.

\paragraph{Views -- Hard tier (10).}
Clean views + cropped at the edge of the frame, distant subject in the background.

\paragraph{Lighting (8).}
Sunny afternoon, overcast daylight, golden hour, indoor soft window light, nighttime under streetlights, rainy with wet reflections, foggy morning, backlit rim light.

\paragraph{Hard modifiers (5).}
Partially occluded by a foreground object, slight motion blur, busy background but still one main subject, in the distance, cropped slightly.

\paragraph{Prompt openers (6).}
A photorealistic color photo, a high-quality image, an ultra-realistic rendering, a detailed photograph, a finely detailed photo, a professional photo.

\subsection{Example Prompts}

\textbf{Clean tier:}
\begin{quote}
\texttt{a photorealistic color photo of a stag with antlers deer standing in a sandy beach, low angle, foggy morning. single main subject, photorealistic color photo, natural colors, realistic texture, sharp focus on subject, no people, no text, no watermark, no logo, no border, no collage, no CGI, no illustration.}
\end{quote}

\textbf{Hard tier:}
\begin{quote}
\texttt{a finely detailed photo of a bullfrog near the ground in a gravel lot, side profile, nighttime under streetlights. partially occluded by a foreground object. single main subject, photorealistic color photo, natural colors, realistic texture, sharp focus on subject, no people, no text, no watermark, no logo, no border, no collage, no CGI, no illustration.}
\end{quote}


\section{Generator Experimental Settings}
\label{app:genconfig}




\subsection{Per-Generator Configuration}

Table~\ref{tab:pergen} details the model-specific inference settings. Key differences include the guidance mechanism (classifier-free guidance scale vs.\ true CFG scale vs.\ flow-based guidance), negative prompt support, and quantization strategy.

\begin{table}[h]
\centering
\caption{Per-generator model configuration and inference settings. The first three generators compose the GenSyn10 training set; SD 3.5 Large is the held-out OOD generator.}
\label{tab:pergen}
\scriptsize
\setlength{\tabcolsep}{2pt}
\begin{tabular}{@{}p{2.2cm}p{2.8cm}p{3.0cm}p{2.8cm}p{2.8cm}@{}}
\toprule
\textbf{Setting} & \textbf{FLUX.2-dev} & \textbf{HunyuanImage-3.0} & \textbf{Qwen-Image-2512} & \textbf{SD 3.5 Large} \\
\midrule
Hub ID & \texttt{black-forest-labs/ FLUX.2-dev} & \texttt{EricRollei/ HunyuanImage-3-NF4-v2} & \texttt{Qwen/ Qwen-Image-2512} & \texttt{stabilityai/ stable-diffusion-
3.5-large} \\
Architecture & Rectified Flow Transformer & MoE Transformer & MMDiT & MMDiT \\
Precision & 4-bit (BnB) + bf16 & NF4 (4-bit) + bf16 & bfloat16 (native) & bfloat16 (native) \\
Quantization & BitsAndBytes 4-bit (transformer + text enc.) & BitsAndBytes NF4 (full model) & None & None \\
Guidance & \texttt{guidance\_scale}=7.5 & \texttt{guidance\_scale}=5.0, \texttt{flow\_shift}=3.0 & \texttt{true\_cfg\_scale}=4.0 & \texttt{guidance\_scale}=7.5 \\
Neg.\ prompt & Not supported & Not used & Supported & Supported \\
Inf.\ steps & 30 & 30 & 30 & 30 \\
\bottomrule
\end{tabular}
\end{table}

\paragraph{FLUX.2-dev.}
The FLUX.2-dev pipeline loads 4-bit quantized weights for both the transformer backbone and the Mistral-3 text encoder from \texttt{diffusers/FLUX.2-dev-bnb-4bit}, with remaining components loaded from the base model ID in bfloat16.
CPU offloading is enabled to reduce peak VRAM usage.
FLUX.2 uses rectified flow matching and does not support negative prompts; the negative prompt field is ignored during inference.
The pipeline call passes \texttt{guidance\_scale=7.5} as the classifier-free guidance parameter.

\paragraph{HunyuanImage-3.0.}
HunyuanImage-3.0 is loaded via \texttt{AutoModelForCausalLM} with \texttt{trust\_remote\_code=True}, which downloads the custom pipeline code from the Hub repository.
The model uses NF4 quantization with automatic device mapping across available GPUs.
Image generation is performed via the model's \texttt{generate\_image()} method with \texttt{stream=True}, passing \texttt{diff\_infer\_steps=30} and \texttt{image\_size=(512, 512)}.
The model internally applies \texttt{guidance\_scale=5.0} and \texttt{flow\_shift=3.0}.

\paragraph{Qwen-Image-2512.}
Qwen-Image-2512 is loaded via \texttt{DiffusionPipeline.from\_pretrained} in native bfloat16 precision (no quantization required).
It is the only generator in GenSyn10 that uses a \texttt{true\_cfg\_scale} parameter (set to 4.0) rather than the standard \texttt{guidance\_scale}, and the only one that actively uses the negative prompt during inference.

\paragraph{Stable Diffusion 3.5 Large (held-out).}
SD~3.5 Large~\cite{sd35} is loaded via \texttt{StableDiffusion3Pipeline.from\_pretrained} in native bfloat16 precision without quantization.
It uses the standard \texttt{guidance\_scale=7.5} and supports negative prompts.
The same prompt grammar, inference steps (30), resolution ($512\!\times\!512$), and Lanczos downsampling pipeline are applied identically to the three training generators, ensuring that the generator architecture is the sole variable differing between in-distribution and OOD evaluation.
SD~3.5 Large shares the MMDiT paradigm with Qwen-Image but differs in its text encoding stack (three text encoders: CLIP~ViT-L, OpenCLIP~ViT-bigG, and T5-XXL, vs.\ Qwen-Image's VLM-based encoder) and internal architecture details.

\subsection{Hardware}

All image generation was performed on an NVIDIA RTX PRO 6000 Blackwell GPU (96\,GB VRAM).
4-bit quantization (BitsAndBytes / NF4) was applied to FLUX.2-dev and HunyuanImage-3.0 to fit within memory constraints.
TF32 matrix multiplication was enabled for all generators (\texttt{torch.backends.cuda.matmul.allow\_tf32 = True}).

\subsection{Image Samples}

Figures~\ref{fig:flux_samples}--\ref{fig:sd35_samples} provide representative samples from the four generators across the 10 CIFAR-10-aligned classes. These examples illustrate the visual diversity, class consistency, and generator-specific stylistic differences present in the synthetic subsets.
\newcommand{\genrow}[2]{%
#1 &
\includegraphics[width=.09\textwidth]{images/sample_generator_images/#2/#1/1.png} &
\includegraphics[width=.09\textwidth]{images/sample_generator_images/#2/#1/2.png} &
\includegraphics[width=.09\textwidth]{images/sample_generator_images/#2/#1/3.png} &
\includegraphics[width=.09\textwidth]{images/sample_generator_images/#2/#1/4.png} &
\includegraphics[width=.09\textwidth]{images/sample_generator_images/#2/#1/5.png} &
\includegraphics[width=.09\textwidth]{images/sample_generator_images/#2/#1/6.png} &
\includegraphics[width=.09\textwidth]{images/sample_generator_images/#2/#1/7.png} &
\includegraphics[width=.09\textwidth]{images/sample_generator_images/#2/#1/8.png} \\[-2pt]
}

\begin{figure}[htbp]
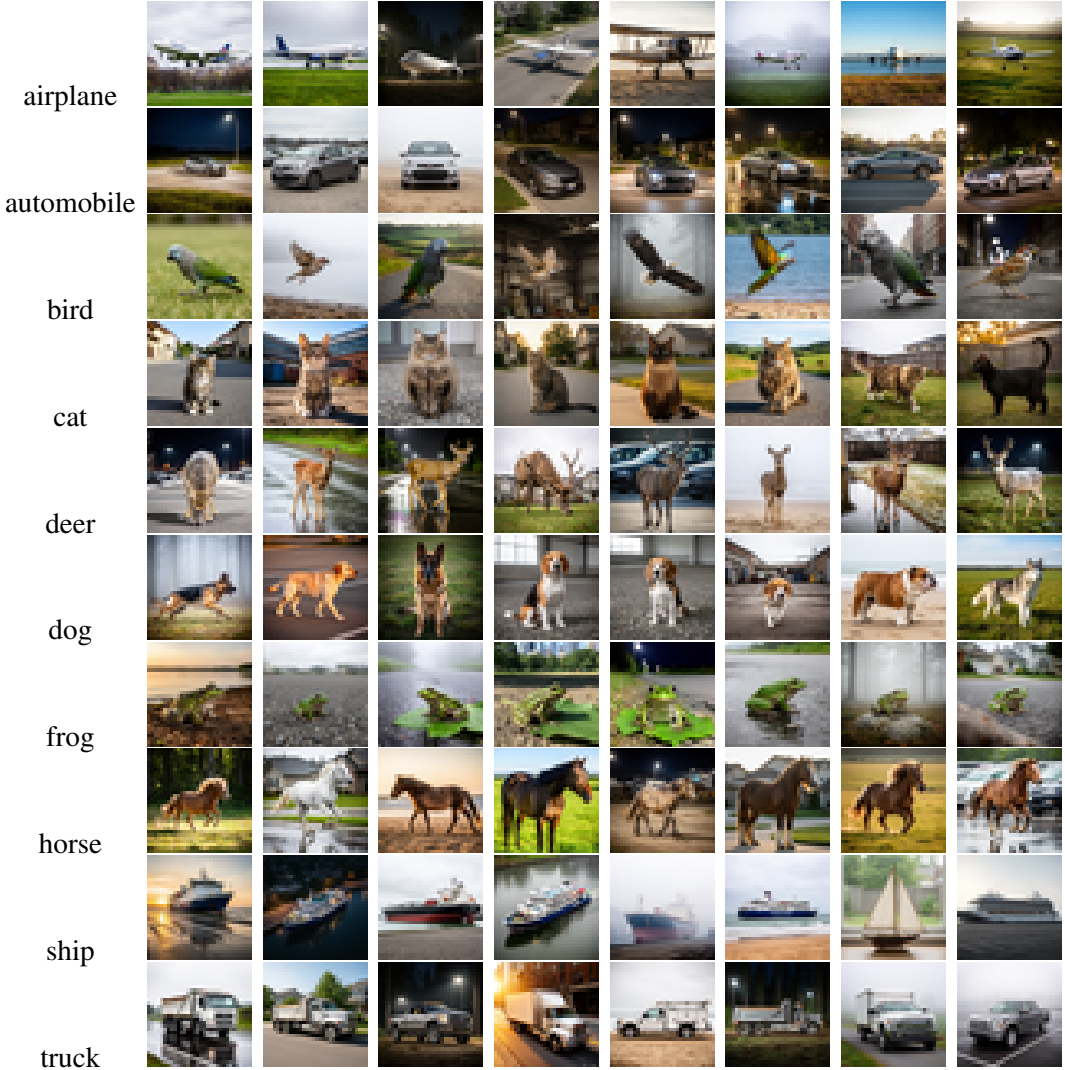

    \centering
    \setlength{\tabcolsep}{2pt}
    \renewcommand{\arraystretch}{0.9}

    \resizebox{\linewidth}{!}{%
    \begin{tabular}{@{}c *{8}{c}@{}}
      & \multicolumn{8}{c}{}\\[-12pt]
      \genrow{airplane}{FLUX}
      \genrow{automobile}{FLUX}
      \genrow{bird}{FLUX}
      \genrow{cat}{FLUX}
      \genrow{deer}{FLUX}
      \genrow{dog}{FLUX}
      \genrow{frog}{FLUX}
      \genrow{horse}{FLUX}
      \genrow{ship}{FLUX}
      \genrow{truck}{FLUX}
    \end{tabular}%
    }

    \caption{Sample images generated by FLUX.2-dev across the 10 CIFAR-10-aligned categories.}
    \label{fig:flux_samples}
\end{figure}

\begin{figure}[htbp]
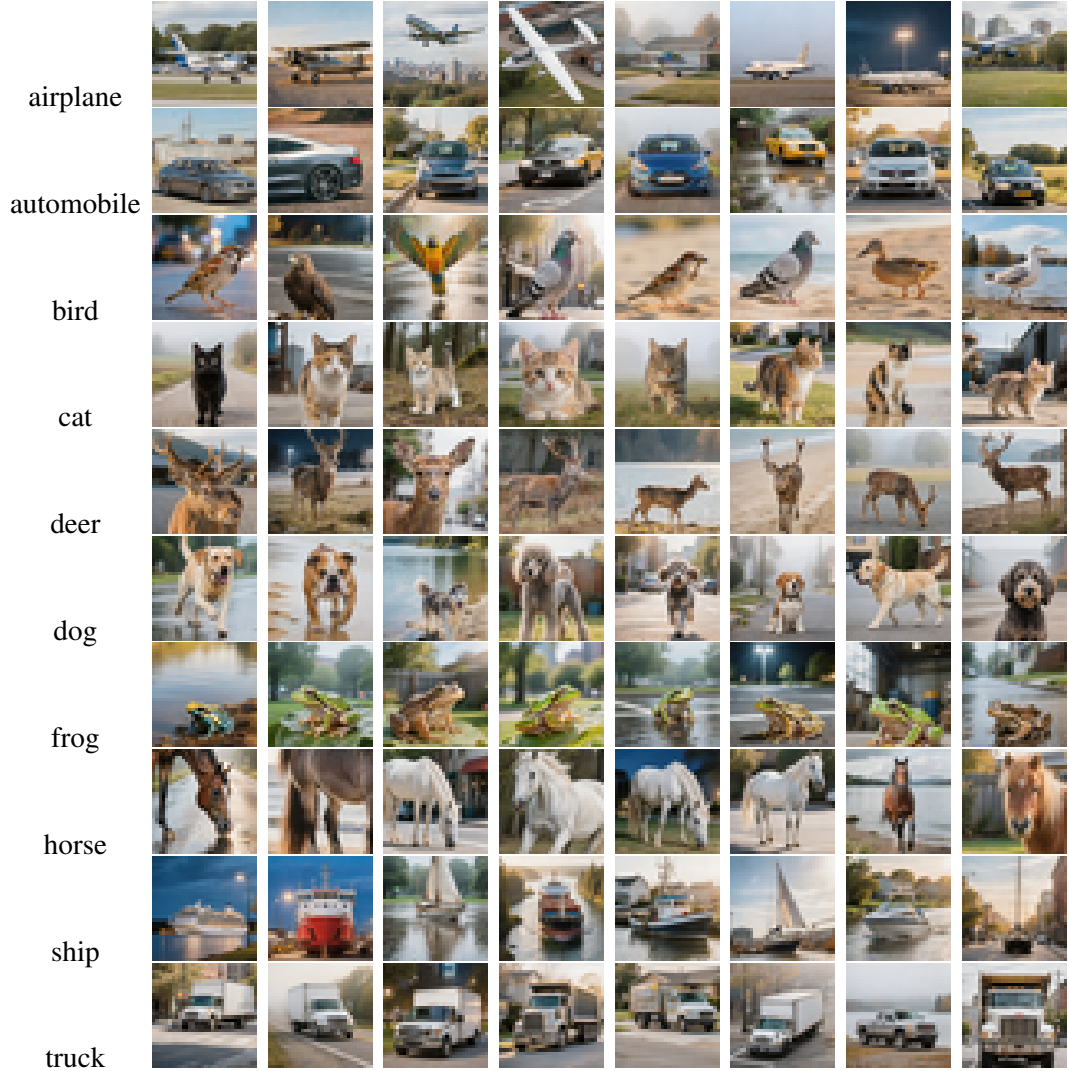

    \centering
    \setlength{\tabcolsep}{2pt}
    \renewcommand{\arraystretch}{0.9}

    \resizebox{\linewidth}{!}{%
    \begin{tabular}{@{}c *{8}{c}@{}}
      \genrow{airplane}{Hunyuan}
      \genrow{automobile}{Hunyuan}
      \genrow{bird}{Hunyuan}
      \genrow{cat}{Hunyuan}
      \genrow{deer}{Hunyuan}
      \genrow{dog}{Hunyuan}
      \genrow{frog}{Hunyuan}
      \genrow{horse}{Hunyuan}
      \genrow{ship}{Hunyuan}
      \genrow{truck}{Hunyuan}
    \end{tabular}%
    }

    \caption{Sample images generated by HunyuanImage-3.0 across the 10 CIFAR-10-aligned categories.}
    \label{fig:hunyuan_samples}
\end{figure}

\begin{figure}[htbp]
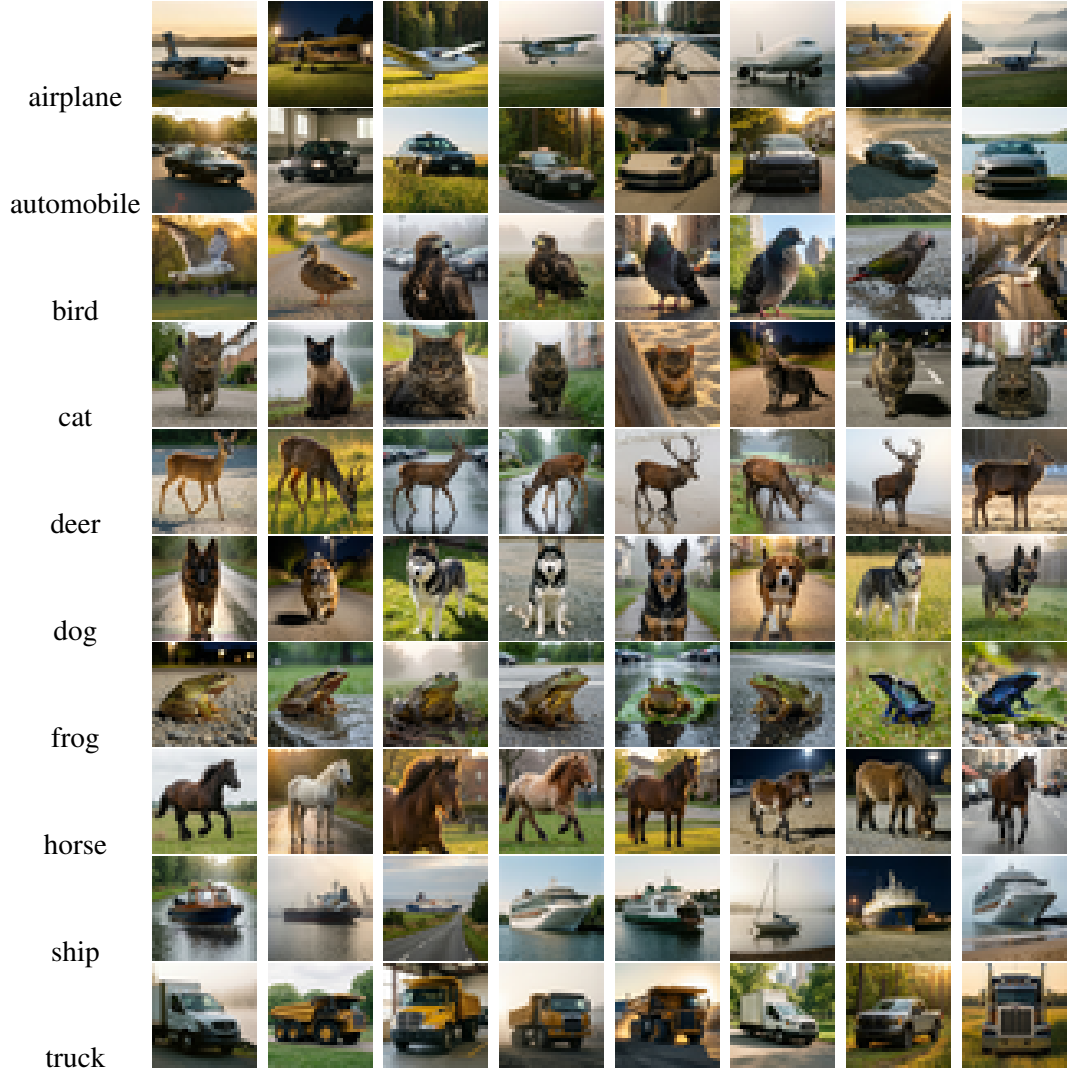

    \centering
    \setlength{\tabcolsep}{2pt}
    \renewcommand{\arraystretch}{0.9}

    \resizebox{\linewidth}{!}{%
    \begin{tabular}{@{}c *{8}{c}@{}}
      \genrow{airplane}{Qwen}
      \genrow{automobile}{Qwen}
      \genrow{bird}{Qwen}
      \genrow{cat}{Qwen}
      \genrow{deer}{Qwen}
      \genrow{dog}{Qwen}
      \genrow{frog}{Qwen}
      \genrow{horse}{Qwen}
      \genrow{ship}{Qwen}
      \genrow{truck}{Qwen}
    \end{tabular}%
    }

    \caption{Sample images generated by Qwen-Image-2512 across the 10 CIFAR-10-aligned categories.}
    \label{fig:qwen_samples}
\end{figure}

\begin{figure}[htbp]
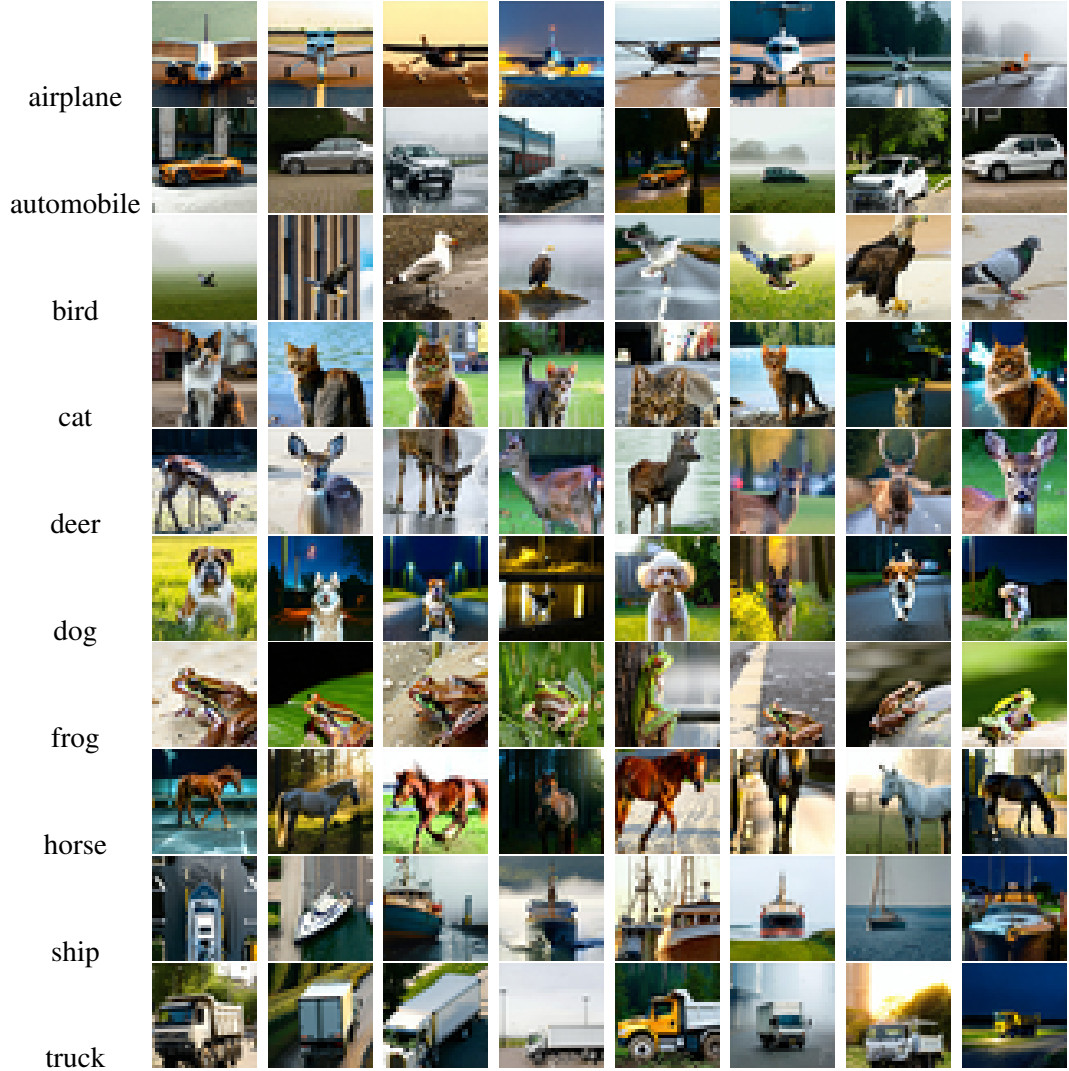

    \centering
    \setlength{\tabcolsep}{2pt}
    \renewcommand{\arraystretch}{0.9}

    \resizebox{\linewidth}{!}{%
    \begin{tabular}{@{}c *{8}{c}@{}}
      \genrow{airplane}{SD}
      \genrow{automobile}{SD}
      \genrow{bird}{SD}
      \genrow{cat}{SD}
      \genrow{deer}{SD}
      \genrow{dog}{SD}
      \genrow{frog}{SD}
      \genrow{horse}{SD}
      \genrow{ship}{SD}
      \genrow{truck}{SD}
    \end{tabular}%
    }

    \caption{Sample images generated by Stable Diffusion 3.5 Large across the 10 CIFAR-10-aligned categories.}
    \label{fig:sd35_samples}
\end{figure}

\section{Classifier Experimental Settings}
\label{app:classifier_settings}

\subsection{Data Loading and Preprocessing}

\paragraph{Input pipelines.}
Three preprocessing pipelines are used depending on the model's expected input resolution:

\begin{table}[h]
\centering
\caption{Preprocessing pipelines for classifier evaluation and training.}
\label{tab:preprocess}
\small
\begin{tabular}{@{}llp{6.5cm}@{}}
\toprule
\textbf{Pipeline} & \textbf{Resolution} & \textbf{Transforms} \\
\midrule
Pipeline A (32$\times$32) & $32\!\times\!32$ & ToTensor $\rightarrow$ Normalize($\mu$=[0.491, 0.482, 0.447], $\sigma$=[0.202, 0.199, 0.201]) \\
Pipeline B (32$\times$32, alt.) & $32\!\times\!32$ & ToTensor $\rightarrow$ Normalize($\mu$=[0.491, 0.482, 0.447], $\sigma$=[0.247, 0.244, 0.262]) \\
Pipeline C (224$\times$224) & $224\!\times\!224$ & Resize(256, bicubic) $\rightarrow$ CenterCrop(224) $\rightarrow$ ToTensor $\rightarrow$ Normalize($\mu$=[0.485, 0.456, 0.406], $\sigma$=[0.229, 0.224, 0.225]) \\
\bottomrule
\end{tabular}
\end{table}

Pipeline A uses the standard CIFAR-10 normalization statistics and is applied to \texttt{chenyaofo/pytorch-cifar-models} checkpoints (VGG, MobileNetV2).
Pipeline B uses the normalization from \texttt{huyvnphan/PyTorch\_CIFAR10} (GoogLeNet, InceptionV3, DenseNets, ResNets).
Pipeline C uses ImageNet normalization and is applied to all models requiring $224\!\times\!224$ input (ViT, Swin, ConvNeXt, EfficientNet, RegNet, MNASNet, and any transfer-learned model).

\paragraph{Training augmentation (Pipeline C only).}
For fine-tuning, training images use: RandomCrop(32, padding=4) $\rightarrow$ Resize(224, bicubic) $\rightarrow$ RandomHorizontalFlip $\rightarrow$ ColorJitter(brightness=0.2, contrast=0.2, saturation=0.2) $\rightarrow$ ToTensor $\rightarrow$ ImageNet Normalize.

\paragraph{Data splits.}
For CIFAR-10 initialization, a stratified 90/10 train/validation split is created from the 50{,}000 CIFAR-10 training images (seed 42), yielding 45{,}000 training and 5{,}000 validation images.
The same strategy is applied to the GenSyn10 training set for GenSyn10 fine-tuning.
All splits preserve class balance.

\subsection{Training Configuration}

\begin{table}[t]
\centering
\caption{Common training hyperparameters for all classifier experiments.}
\label{tab:train_common}
\small
\begin{tabular}{@{}ll@{}}
\toprule
\textbf{Parameter} & \textbf{Value} \\
\midrule
Optimizer & AdamW \\
Loss function & CrossEntropyLoss with label smoothing ($\epsilon = 0.05$) \\
LR scheduler & ReduceLROnPlateau (mode=max, factor=0.5, scheduler patience=1, min\_lr=$10^{-6}$) \\
Gradient clipping & Max norm = 1.0 \\
Mixed precision & AMP (automatic mixed precision) on CUDA \\
Batch size & 200 (CUDA) / 64 (CPU) \\
Checkpoint selection & Composite: $0.7 \times \text{val\_macro\_F1} + 0.3 \times \text{val\_balanced\_accuracy}$ \\
Random seeds & torch.manual\_seed(42), numpy.random.seed(42) \\
\bottomrule
\end{tabular}
\end{table}

\noindent Family-specific hyperparameters (head epochs, full epochs, learning rates, weight decay, early-stopping patience) are detailed in Table~\ref{tab:ftprotocol}.

\begin{table}[t]
\centering
\caption{Family-aware fine-tuning hyperparameters. Transformer and hybrid architectures use more conservative learning rates, longer training budgets, and higher patience thresholds than generic CNNs to reduce catastrophic forgetting.}
\label{tab:ftprotocol}
\small
\setlength{\tabcolsep}{4pt}
\renewcommand{\arraystretch}{1.15}
\begin{tabular}{@{}p{3.4cm}cccccc@{}}
\toprule
\textbf{Family} &
\textbf{\shortstack{Head\\Epochs}} &
\textbf{\shortstack{Full\\Epochs}} &
$\mathbf{LR}_{\textbf{head}}$ &
$\mathbf{LR}_{\textbf{full}}$ &
\textbf{\shortstack{Weight\\Decay}} &
\textbf{Patience} \\
\midrule
Generic CNNs (VGG, ResNet, DenseNet, Inception)
& 2 & 8  & $10^{-3}$            & $10^{-4}$            & $5 \times 10^{-5}$ & 3 \\
MNASNet
& 3 & 10 & $10^{-3}$            & $8 \times 10^{-5}$   & $10^{-4}$          & 4 \\
ViT, Swin
& 3 & 12 & $8 \times 10^{-4}$   & $5 \times 10^{-5}$   & $10^{-4}$          & 4 \\
ConvNeXt, ConvNeXtV2
& 3 & 12 & $10^{-3}$            & $7 \times 10^{-5}$   & $10^{-4}$          & 4 \\
\bottomrule
\end{tabular}
\end{table}

\subsection{Evaluation Metrics}

All experiments report the following metrics, computed via \texttt{scikit-learn}:

\begin{table}[t]
\centering
\caption{Evaluation metrics reported in the paper. The first group is shared across all experiments; the second group is reported only for the binary real-vs-synthetic detection task.}
\label{tab:metrics_list}
\small
\begin{tabular}{@{}llp{6.5cm}@{}}
\toprule
\textbf{Scope} & \textbf{Metric} & \textbf{Description} \\
\midrule
\multirow{4}{*}{All experiments} 
 & Balanced Accuracy & Macro-averaged per-class accuracy \\
 & F1 Score & Macro-averaged F1 \\
 & MCC & Matthews Correlation Coefficient \\
\midrule
\multirow{4}{*}{Binary only} & ROC AUC & Area under the ROC curve \\
 & PR AUC & Area under the precision--recall curve \\
 & Log Loss & Negative log-likelihood of predicted probabilities \\
 & Brier Score & Mean squared error of predicted probabilities \\
\bottomrule
\end{tabular}
\end{table}

\noindent Probability outputs are obtained from softmax over model logits.

\section{Full Four-Stage Results with Additional Metrics}
\label{app:fourstage_full}

\begin{figure}[tbp]
  \centering
  \includegraphics[width=1\textwidth]{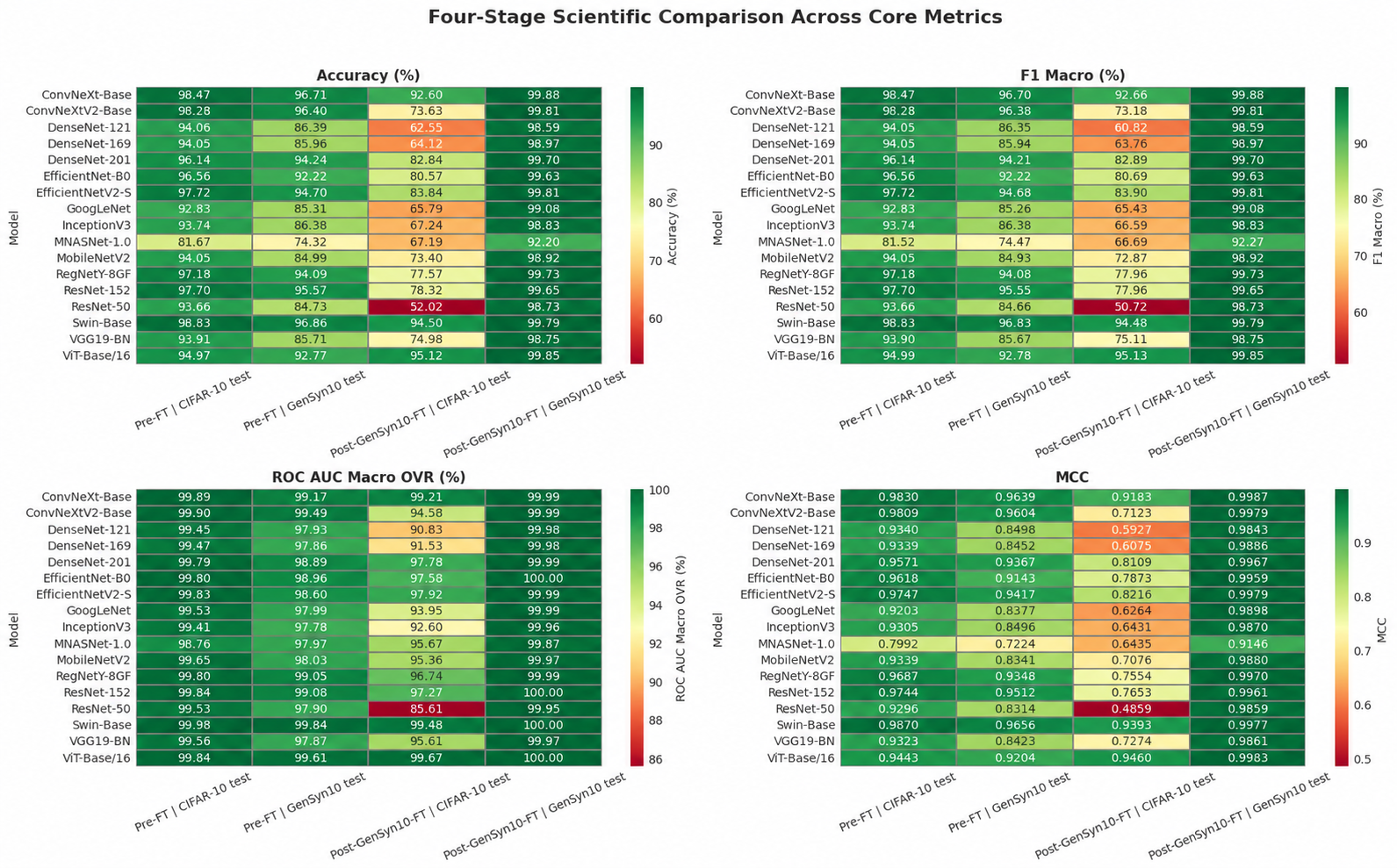}
  \caption{Four-stage comparison across core metrics.}
  \label{fig:fourstage_full}
\end{figure}

Figure~\ref{fig:fourstage_full} extends Table~\ref{tab:fourstage} ROC AUC, and MCC for each stage.

\section{GenSynCIFAR10 Dataset Structure}
\label{app:GenSynCIFAR10}

The GenSynCIFAR10 binary classification dataset is structured as follows:

\begin{verbatim}
GenSynCIFAR10/
+-- train/
|   +-- real/       # CIFAR-10 training images
|   +-- synthetic/  # GenSyn10 training images
+-- test/
    +-- real/       # CIFAR-10 test images
    +-- synthetic/  # GenSyn10 test images
\end{verbatim}

\noindent The train split contains 50{,}000 real + 50{,}000 synthetic images.
The test split contains 10{,}000 real + 10{,}000 synthetic images.
For OOD evaluation, an additional test set replaces the synthetic images with outputs from Stable Diffusion 3.5 Large, generated using the same prompt grammar and downsampling pipeline.

\section{Binary Classification: Full Ranking}
\label{app:binary_ranking}

Tables~\ref{tab:binary_ranking} and~\ref{tab:binary_ranking_ood} present the complete rankings of all 17 architectures on the GenSynCIFAR10 binary detection task for in-distribution (seen generators) and out-of-distribution (SD 3.5 Large) evaluation, respectively.
Models are ranked by mean rank across all metrics; ties are broken by accuracy.

\begin{table}[t]
\centering
\caption{In-distribution binary detection ranking on GenSynCIFAR10 (seen generators).
Models are ordered by mean rank across all metrics.
Acc = accuracy (\%), F1 = macro F1 (\%), MCC = Matthews Correlation Coefficient.
Best values per column are \textbf{bolded}.}
\label{tab:binary_ranking}
\scriptsize
\setlength{\tabcolsep}{2.5pt}
\begin{tabular}{@{}cl ccccc cc c@{}}
\toprule
\textbf{Rank} & \textbf{Model} & \textbf{Acc} & \textbf{F1} & \textbf{MCC} & \textbf{ROC AUC} & \textbf{PR AUC} & \textbf{Log Loss} & \textbf{Brier} & \textbf{Mean Rank} \\
\midrule
1  & Swin-Base        & \textbf{99.94} & \textbf{99.94} & \textbf{0.999} & \textbf{100.00} & \textbf{100.00} & \textbf{0.015} & \textbf{0.001} & 1.4 \\
2  & ViT-Base/16      & 99.88 & 99.88 & 0.998 & \textbf{100.00} & \textbf{100.00} & 0.020 & 0.001 & 3.2 \\
3  & ConvNeXt-Base    & 99.91 & 99.91 & 0.998 & 99.99 & 99.99 & 0.017 & 0.001 & 3.8 \\
4  & ResNet-152       & 99.77 & 99.77 & 0.995 & \textbf{100.00} & \textbf{100.00} & 0.023 & 0.002 & 4.7 \\
5  & DenseNet-201     & 99.69 & 99.69 & 0.994 & \textbf{100.00} & \textbf{100.00} & 0.018 & 0.003 & 5.4 \\
6  & EfficientNetV2-S & 99.72 & 99.71 & 0.994 & 99.99 & \textbf{100.00} & 0.023 & 0.003 & 5.5 \\
7  & GoogLeNet        & 99.78 & 99.78 & 0.996 & 99.99 & 99.98 & 0.021 & 0.002 & 5.8 \\
8  & RegNetY-8GF      & 99.59 & 99.59 & 0.992 & \textbf{100.00} & \textbf{100.00} & 0.027 & 0.003 & 7.4 \\
9  & MobileNetV2      & 99.61 & 99.61 & 0.992 & 99.97 & 99.95 & 0.023 & 0.003 & 9.6 \\
10 & ResNet-50        & 99.59 & 99.59 & 0.992 & 99.98 & 99.98 & 0.026 & 0.004 & 9.8 \\
11 & DenseNet-121     & 99.50 & 99.50 & 0.990 & 99.99 & 99.99 & 0.028 & 0.004 & 10.9 \\
12 & VGG19-BN         & 99.52 & 99.52 & 0.990 & 99.97 & 99.96 & 0.031 & 0.004 & 11.7 \\
13 & ConvNeXtV2-Base  & 97.34 & 97.27 & 0.948 & \textbf{100.00} & \textbf{100.00} & 0.098 & 0.023 & 12.3 \\
14 & DenseNet-169     & 99.33 & 99.33 & 0.987 & 99.97 & 99.97 & 0.032 & 0.005 & 13.1 \\
15 & EfficientNet-B0  & 99.07 & 99.06 & 0.982 & 99.97 & 99.98 & 0.039 & 0.008 & 13.1 \\
16 & InceptionV3      & 99.31 & 99.31 & 0.986 & 99.96 & 99.95 & 0.035 & 0.006 & 13.7 \\
17 & MNASNet-1.0      & 97.70 & 97.66 & 0.955 & 99.96 & 99.95 & 0.063 & 0.018 & 15.3 \\
\bottomrule
\end{tabular}
\end{table}

\begin{table}[h]
\centering
\caption{OOD binary detection ranking on Stable Diffusion 3.5 Large (unseen generator).
Models are ordered by mean rank across all metrics.
Acc = accuracy (\%), F1 = macro F1 (\%), MCC = Matthews Correlation Coefficient.
Best values per column are \textbf{bolded}.}
\label{tab:binary_ranking_ood}
\scriptsize
\setlength{\tabcolsep}{2.5pt}
\begin{tabular}{@{}cl ccccc cc c@{}}
\toprule
\textbf{Rank} & \textbf{Model} & \textbf{Acc} & \textbf{F1} & \textbf{MCC} & \textbf{ROC AUC} & \textbf{PR AUC} & \textbf{Log Loss} & \textbf{Brier} & \textbf{Mean Rank} \\
\midrule
1  & ViT-Base/16      & \textbf{96.08} & \textbf{95.92} & \textbf{0.924} & 99.78 & 99.83 & \textbf{0.111} & \textbf{0.030} & 2.2 \\
2  & Swin-Base        & 95.54 & 95.33 & 0.914 & 99.66 & 99.77 & 0.134 & 0.036 & 3.2 \\
3  & ConvNeXt-Base    & 94.71 & 94.42 & 0.899 & \textbf{99.87} & \textbf{99.89} & 0.157 & 0.042 & 4.6 \\
4  & InceptionV3      & 95.36 & 95.18 & 0.909 & 99.49 & 99.50 & 0.131 & 0.036 & 4.9 \\
5  & GoogLeNet        & 94.87 & 94.61 & 0.902 & 98.89 & 99.31 & 0.152 & 0.041 & 6.6 \\
6  & DenseNet-201     & 91.69 & 90.94 & 0.845 & 99.62 & 99.71 & 0.235 & 0.066 & 7.7 \\
7  & DenseNet-169     & 92.89 & 92.41 & 0.865 & 99.35 & 99.38 & 0.200 & 0.056 & 8.6 \\
8  & RegNetY-8GF      & 90.97 & 90.08 & 0.833 & 99.71 & 99.78 & 0.230 & 0.069 & 8.7 \\
9  & MobileNetV2      & 93.23 & 92.76 & 0.872 & 98.29 & 98.92 & 0.202 & 0.054 & 8.9 \\
10 & ResNet-152       & 91.32 & 90.50 & 0.839 & 99.20 & 99.48 & 0.231 & 0.066 & 9.0 \\
11 & EfficientNetV2-S & 91.27 & 90.43 & 0.838 & 98.85 & 99.28 & 0.241 & 0.069 & 10.3 \\
12 & VGG19-BN         & 90.75 & 89.86 & 0.828 & 99.42 & 99.43 & 0.204 & 0.062 & 11.0 \\
13 & ConvNeXtV2-Base  & 79.19 & 73.72 & 0.642 & 99.88 & 99.88 & 0.760 & 0.191 & 12.4 \\
14 & EfficientNet-B0  & 86.62 & 84.57 & 0.760 & 99.29 & 99.47 & 0.344 & 0.103 & 13.0 \\
15 & MNASNet-1.0      & 85.48 & 83.06 & 0.741 & 99.49 & 99.48 & 0.488 & 0.117 & 13.7 \\
16 & ResNet-50        & 89.85 & 88.76 & 0.812 & 98.41 & 98.80 & 0.305 & 0.084 & 13.8 \\
17 & DenseNet-121     & 89.64 & 88.48 & 0.809 & 98.85 & 99.14 & 0.312 & 0.085 & 13.8 \\
\bottomrule
\end{tabular}
\end{table}


\clearpage

\newpage

\end{document}